\documentclass[]{fairmeta}

\usepackage{hyperref}
\usepackage{url}
\usepackage{algorithm}
\usepackage{algpseudocode}
\usepackage{xcolor}

\definecolor{mygray}{gray}{0.95}

\usepackage{amsmath,amsfonts,bm}

\def\eqref#1{equation~\ref{#1}}

\def\1{\bm{1}}

\def\vzero{{\bm{0}}}

\def\vmu{{\bm{\mu}}}

\def\ve{{\bm{e}}}

\def\vx{{\bm{x}}}
\def\vy{{\bm{y}}}
\def\vz{{\bm{z}}}

\def\vsigma{{\bm{\sigma}}}

\DeclareMathAlphabet{\mathsfit}{\encodingdefault}{\sfdefault}{m}{sl}
\SetMathAlphabet{\mathsfit}{bold}{\encodingdefault}{\sfdefault}{bx}{n}

\def\gN{{\mathcal{N}}}

\def\sS{{\mathbb{S}}}

\newcommand{\E}{\mathbb{E}}

\usepackage{tabularx,ragged2e,booktabs,caption}
\newcolumntype{C}[1]{>{\Centering}m{#1}}

\newcolumntype{Z}[1]{>{\Left}m{#1}}

\title{A Unified Knowledge-Distillation and Semi-Supervised Learning Framework to Improve Industrial Ads Delivery Systems}

\author[1]{Hamid Eghbalzadeh, Yang Wang, Rui Li, Yuji Mo, Qin Ding, Jiaxiang Fu, Liang Dai, Shuo Gu, Nima Noorshams, Sem Park, Bo Long, Xue Feng}

\affiliation[1]{AI at Meta}

\abstract{
Industrial ads ranking systems conventionally rely on labeled impression data, which leads to challenges such as overfitting, slower incremental gain from model scaling, and biases due to discrepancies between training and serving data. 
To overcome these issues, we propose a Unified framework for Knowledge-Distillation and Semi-supervised Learning (UKDSL) for ads ranking, empowering the training of models on a significantly larger and more diverse datasets, thereby reducing overfitting and mitigating training-serving data discrepancies. 
We provide detailed formal analysis and numerical simulations on the inherent miscalibration and prediction bias of multi-stage ranking systems, and show empirical evidence of the proposed framework's capability to mitigate those.  Compared to prior work, UKDSL can enable models to learn from a much larger set of unlabeled data, hence, improving the performance while being computationally efficient. 
Finally, we report the successful deployment of UKDSL in an industrial setting across various ranking models, serving users at multi-billion scale, across various surfaces, geological locations, clients, and optimize for various events, which to the best of our knowledge is the first of its kind in terms of the scale and efficiency at which it operates. 
}

\date{2024-12-12}
\correspondence{Xue Feng at \email{xfeng@meta.com}}

\begin{document}

\maketitle

\section{Introduction}
\begin{figure}
\centering
\includegraphics[width=7cm]{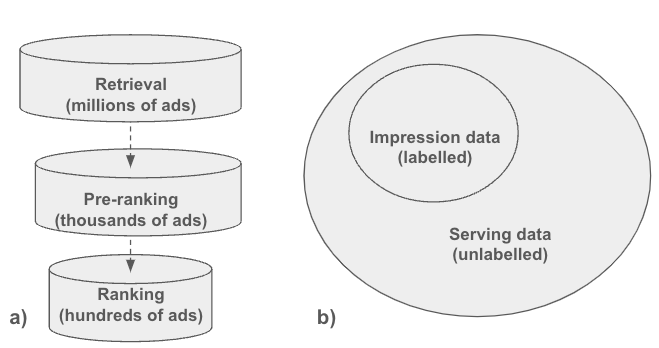}
  \caption{a) a block-diagram of a multi-stage ranking system. b) demonstration of unlabeled and labeled data used in industrial systems.}
  \label{fig:teaser}
\end{figure}
When users surf on applications with large amounts of users (e.g., social media, streaming services, online shops, etc), they would like to see ads that they find relevant to their interests and needs. To achieve this goal, ads ranking systems are built to find a handful of best matched ads from the candidate pool of large number of ads within seconds.

To balance the accuracy and efficiency under such tight time and large scale constraints, industrial ads ranking systems use a cascade multi-stage ranking design~\cite{wang2023towards}. Each stage is responsible for scanning through the candidates provided by the previous stage, and provides the top candidates for the next stage (see Figure 1). Typically, the models used in earlier stages (e.g., retrieval stage) are simpler, more efficient but less accurate (e.g., two tower network~\cite{scalingIG23}), whereas the models used in later stages are more complex, accurate, but slower. After the top ads were delivered to the users, feedback from the users (e.g., click) on those delivered ads (i.e. impression data) can be collected as labels. 

Traditionally, the models from all stages are trained with the impression data in a supervised fashion. However, the impression data is only a very small subset of the data that the early stage models (i.e., retrieval and pre-ranking stage models) see and make predictions about. Hence, there are large discrepancies between training and serving data for the early stage models. Such discrepancies are often a cause for online performance degradations due to lack of generalization.

In this work, we leverage a Unified framework for Knowledge-Distillation and Semi-supervised Learning (UKDSL) to improve models by closing the gap between the training and serving data. We evaluate the UKDSL framework on industry-scale ads ranking systems across different stages (retrieval and pre-ranking stages), different types (click-through rate models and conversion rate models).
In summary, the contributions of our paper are as follows:
\newline
\begin{itemize}
    \item \textbf{Inherent drawbacks of multi-stage ranking systems}: We provide theoretical analyses revealing  two important inherent problems in multi-stage ranking models. Through formal analysis and numerical simulations, we shed light on causes of these drawbacks.
    \item \textbf{\textbf{U}nified framework for Knowledge-Distillation and Semi-Supervised Learning}: We present \textbf{UKDSL}, a \textbf{U}nified framework for \textbf{K}nowledge-\textbf{D}istillation and \textbf{S}emi-supervised \textbf{L}earning in industrial ads delivery systems. We provide strong empirical evidence of the effectiveness of this framework in improving both offline and online ads rankings performance. We present empirical evidence on applying the UKDSL framework on industry-scale ads ranking systems, showing this method can bring significant improvement to the ads ranking systems.    
\end{itemize}
\vspace{6pt}
The remainder of this paper is as follows. In Section~\ref{sec:related} we present related work and existing methods from the literature.
Section~\ref{sec:problem} will present the problem statement, definitions, and theoretical results.
We detail our UKDSL framework in Section~\ref{sec:method} and present our empirical evidence.
And finally, Section~\ref{sec:discussions_future} concludes this work and portray possible future directions for the research community.

\section{Related Work}
\label{sec:related}
\textbf{Multi-stage ranking} A multi-stage ranking framework is widely adopted within an industrial recommendation system in order to balance between the efficiency and accuracy \cite{deepRSsurvey}. Covington et al. \cite{youtube} employed deep neural network models for both candidate generate stage and ranking stage for Youtube recommendation, Huang et al. \cite{fbsearchemb} designed the Facebook search system with three stages (indexing, retrieval and ranking) under a unified embedding based framework. Six different stages including retrieval, pre-ranking, relevance ranking, ranking, re-ranking, mix-ranking were deployed to build a search system as described in \cite{taobaosearch}.  \\

\noindent \textbf{Sample selection bias} Within a multi-stage ranking system, it has been shown that sample selection bias may be induced in earlier stages, if the data related to delivered items was used to train the early stage models~\cite{pin2024kdd}. There have been different types of approaches to mitigate this bias in the literature. For instance, authors in \cite{pin2024kdd} proposed two variants of unsupervised domain adaptation, and have associated a manually chosen threshold for hard labeling. 
Qin el, al. \cite{rankflow} trained different stages in a cascading flow fashion by training each stage separately followed by cross stage distillations. Building on top of \cite{rankflow}, Zhao et, al.\cite{COPR} emphasized the importance of rank order alignment between stages with bid information, and designed a chunk based data sampling schema to facilitate learning this order alignment. Zheng et al. \cite{zheng2022multiobjective} tackled this problem from a multi-objective perspective by treating samples at by different stages as different class of labels for the model to learn the preference order among relevance, exposure, click and purchase. 
 
Another line of research is around finding or creating new data sources to break the data-model feedback loop within an existing system. Both \cite{rec4ad} and \cite{zhang2023rethinkingsearch} leveraged user interactions from other similar and related scenarios to provide proxy labels for the samples considered by the early stages in the target scenario. \cite{autodebias} and \cite{uniform2020} used a small fraction of traffic to collect uniformly distributed data through random policy.  \\

\noindent \textbf{Knowledge distillation} Knowledge distillation \cite{hinton2015distilling} is a technique for transferring knowledge from a teacher model to a student model with the goal of improving the performance of the student model. A typical setup is that the teacher model is more complex and accurate while the student model is simpler and less accurate. Initially this technique has been applied to domains like computer vision \cite{ba2014l2} \cite{romero2015fitnets} and neural language processing \cite{kim2016sequencelevel}. Later recommendation system started to adopt this technique as well. Tang el al. \cite{rankingdistill} proposed to use the knowledge distillation method to compress their ranking models which had an inference latency constraint while in the serving the traffic. The predictions from multiple teachers are ensembled together through adaptive gating in \cite{ensembledistill} to further boost the performance gain of the distilled student model.

\section{Problem Statement}
\label{sec:problem}

In multi-stage ads ranking systems, one of the major issues is the inconsistency and miscalibration between early and late-stage models.
In this section, we first analytically show how bias is introduced in ranked items, despite the use of unbiased ranking estimators.
Further, we analyze model calibration in various stages via simulation.
Through this, we uncover inherent miscalibrations in such systems.
We then hypothesize a potential solution to this problem which we empirically validate in Section~\ref{sec:method}.

\subsection{Introduction of Bias in Ranked Items via an Unbiased Model}
\label{subsec:introduced_bias}

In this section, we start by analyzing a simple case of one-stage ranking.
Suppose we have $n$ ads candidates with the underlying ground truth Cost Per Mille (CPM) following the distribution of $\vy_i\sim\gN(\vmu, \vsigma^2)$. Let $\vz_i$ be a random variable representing the CPM predictions by some model (here we are simplifying the problem by considering a model outputting the whole eCPM while in reality eCPM was constructed by multiple models). If we assume $\mathbf{M_1}$ to be an unbiased predictor with the prediction error $\ve_1\sim\gN(\vzero, {\vsigma_1}^2)$, we will have $\vz_i\sim\gN(\vmu, \vsigma^2+{\vsigma_1}^2)$. The true value of $\vy_i$ is unknown and only model predictions $\vz_i$ can be observed. Under a typical ranking setup, the model will select the top-k ads based on its own predictions. Let $\sS_1$ be the top-k predicted ads subset selected by $\mathbf{M_1}$ from $n$ available ads, such that  $\sS_1=\{\vz_1, \vz_2, \dots, \vz_k\}$ where $\vz_1>\vz_2>\dots> \vz_k$, . 
We can obtain the expectations of eCPMs following the method described in ~\cite{royston1982expected}:
\begin{align}
    \label{eq:expeCPM}
    \E(\vz_i| \vmu, \vsigma^2,{\vsigma_1}^2,n)=\vmu+
    \sqrt{\vsigma^2+{\vsigma_1}^2}\Phi^{-1}\left(\frac{n-i-\alpha+1}{n-2\alpha+1}  \right)
\end{align}
where $\alpha=3.375$, $\Phi$ is the Gaussian CDF.
With Eq.\ref{eq:expeCPM} in hand, we can now do some analysis on the relationship between eCPMs and CPMs of the selected ads. The goal of this ranking system is to maximize the total return which is the sum of the CPMs of the selected ads. If the model is perfect without any errors, the system reaches its optimal state by finding the top-k CPM ads from the given n ads. The optimal return can be also calculated via Eq.\ref{eq:expeCPM} by setting $\vsigma_1=0$. From Eq.\ref{eq:expeCPM} we can observe that the estimation of the $i^{th}$ ranked item increases as variance of the predictor becomes larger while the total return becomes smaller. In other words, the model predictions are over-calibrated on the set of selected ads. With this toy example, we can see that an unbiased estimator can produce a biased result under a ranking system context. This phenomenon can be understood intuitively by the following argument:
{Although the model's overall prediction is unbiased on the entire candidate set, it is under-calibrated on some parts of the population and over-calibrated on other parts of the population. Therefore, when the model selects the top-k ads based on its prediction, it naturally tends to pick the ones that are over-calibrated. 

\subsection{The inherent Miscalibration in Multi-stage Ranking}
\label{subsec:inherent_miscalibration}
In this section we extend our analysis to a two stage ranking system which better resembles a real world multi-stage ranking system. Suppose we have a first-stage unbiased predictor $\mathbf{M_1}$ with prediction error of $\ve_1\sim\gN(\vzero, {\vsigma_1}^2)$, and a second-stage unbiased predictor $\mathbf{M_2}$ with prediction error of $\ve_2\sim\gN(\vzero, {\vsigma_2}^2)$.
Given the common assumption that early-stage models have higher variance than later stage models, let us further assume that ${\vsigma_1}^2>{\vsigma_2}^2$. Like the discussion in the previous section, we assume we have n ad candidates following the distribution of $\vy_i\sim\gN(\vmu, \vsigma^2)$. The first stage model will select top $k_1$ ads and send them to the second stage, and the second stage model will select top $k_2$ ads and send them to the users. 

The model calibration on the set of ads selected by the first stage ($\sS_1$) is defined as:
\begin{equation}
    \text{cal(i,j)}=\frac{\vmu'_i}{\vmu'_j}
\end{equation}
where ${\vmu'}_i$ and $\vmu'_j$ are average CPM estimation of stages $i$ and $j$ on $\sS_1$, respectively. We use this quantity as an example to illustrate the inherent miscalibration that exist in a multi-stage ranking system. The cross-stage calibration ($\mathbf{M_1}$ w.r.t $\mathbf{M_2}$) is calculated by $cal(1,2)$.
Similarly, we can calculate the calibration of each stage model e.g, $\mathbf{M_j}$ via $cal(j,0)$ where ${\vmu'}_0$ is the average of true CPMs on $\sS_1$.  

In order to analyze the calibration as a function of number of retrieved samples and model characteristics, we run a simulation and show the results in Figure~\ref{fig:two_stage_sim}.
Here, we depict the result of simulation for cross-stage calibration, as well as for each stage's calibration as a function of $k_1$ and different variants of model noise level. 
Regardless of the choice of parameters, we can see that on $\sS_1$ the first stage model is over-calibrated (Figure~\ref{fig:two_stage_sim}-a,b) while the second stage model is well calibrated (Figure~\ref{fig:two_stage_sim}-c). This observation aligns with the intuition discussed in Section ~\ref{subsec:introduced_bias} in that an unbiased model only becomes biased on the set of candidates selected by itself. This simulation result also aligns well with the experimental observation described in Section ~\ref{subsec:semi_consistency_calib}, and at the same time suggests the direction of using the second stage model to correct the bias of the first stage model.

\begin{figure}[h]
  \centering
   \includegraphics[width=8cm]{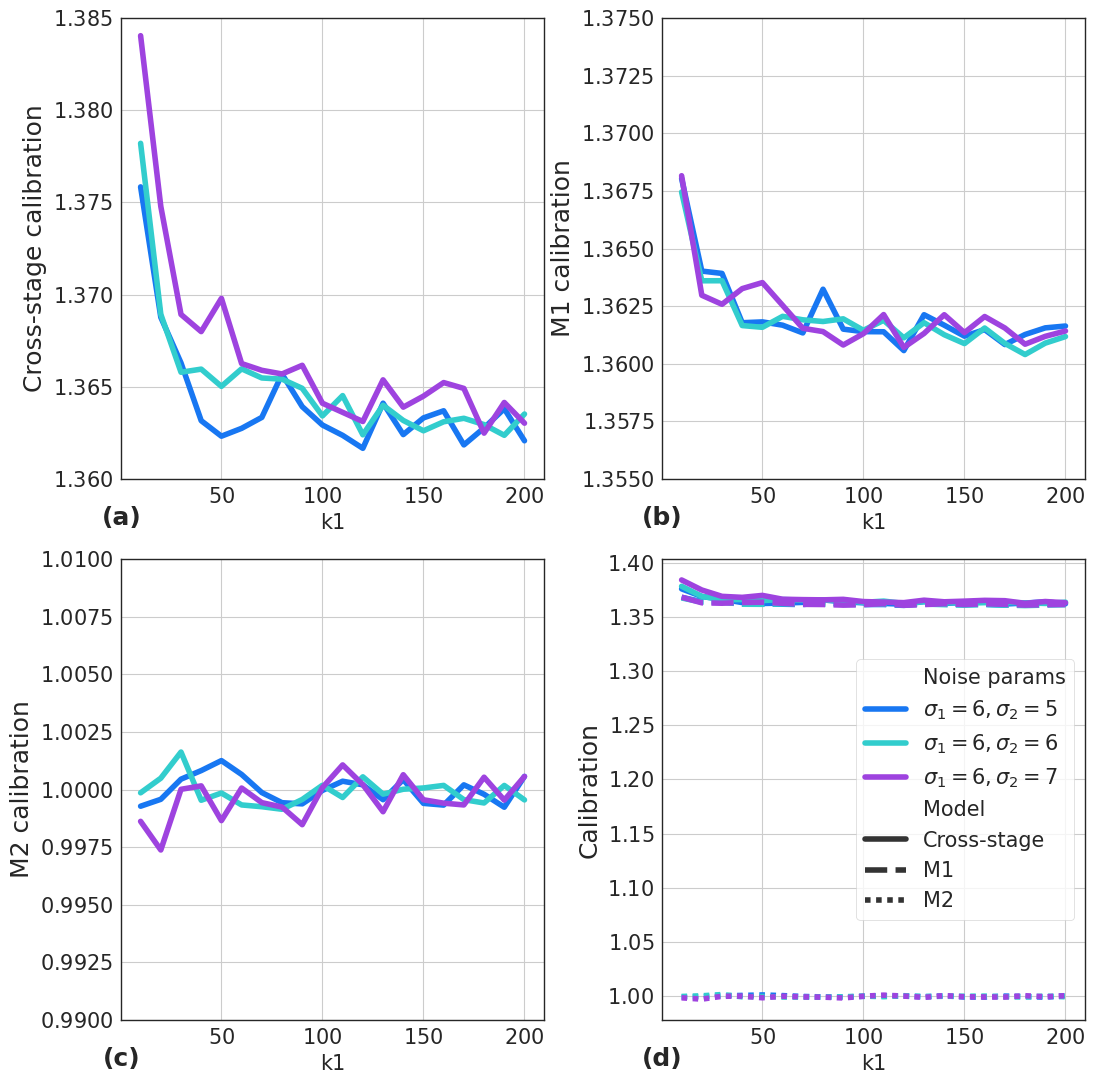}
  \caption{Simulation results for model calibrations. a) $\text{cal}(1,2)$. b)  $\text{cal}(1,0)$. c) $\text{cal}(2,0)$. d) all previous plots along side each other.
  Notes: $\mathbf{M_0}$ denotes the ground truth. All calibrations are calculated on the top $k_1$ ads candidates selected by the first stage model ($\sS_1$).
  }
 \label{fig:two_stage_sim}
\end{figure}

\subsection{Tackling Miscalibration and Introduced Bias}
In Sections \ref{subsec:introduced_bias} and  \ref{subsec:inherent_miscalibration} we presented two important and fundamental problems that are inherent to multi-stage ranking systems.
In the next section (Section~\ref{sec:method}), we investigate solutions to these two problems via \textit{A Unified Framework for Knowledge Distillation and Semi-Supervised Learning (UKDSL)}.
We present empirical evidence in Section~\ref{subsec:semi_consistency_calib} that shows UKDSL improves calibration, hence addressing the first issue discussed in Section~\ref{subsec:inherent_miscalibration}.
Furthermore, we present our empirical evidence on improving later-stage model's performance via leveraging the foundation models~\cite{bommasani2021opportunities}, which have better performance (e.g, lower prediction variance), hence, according to Eq.~\ref{eq:expeCPM}, will decrease the inherent bias of ranking models.

\section{A Unified Framework for Knowledge Distillation and Semi-Supervised Learning (UKDSL)}
\label{sec:method}

In this section, we introduce UKDSL, a Unified Framework for Knowledge Distillation and Semi-Supervised Learning of industrial scale ranking systems, that has been successfully deployed in multi-billion scale industrial settings across various ranking models, serving users at multi-billion scale, across various surfaces, geological locations, clients, and optimize for various events, which to the best of our knowledge is the first of its kind in terms of the scale and efficiency at which it operates. 
In the following, we start by the motivation of the framework, and further detail various modules of UKDSL.

Semi-Supervised Learning (SSL) techniques have been proven useful for mitigating the distribution gap between training and serving data, as discussed in Section~\ref{sec:related}. 
However, for industrial systems, it is difficult to applied SSL techniques at scale. In our attempts to adopt SSL in industrial-scale systems, we face the following 4 variety and scale challenges:
\newline

\begin{enumerate}
    \item Scale and variety of data 
    \item Scale and variety of model types
    \item Variety of ranking stages
    \item Scale and variety of features
\end{enumerate}
\vspace{6pt}
To tackle these challenges, we proposed the Unified Framework for Knowledge Distillation and Semi-Supervised Learning (UKDSL), a framework that's designed to be flexible and scalable for almost all scenarios in a modern ads delivery system with billions of users. UKDSL has 3 main components: (1) Cross-stage knowledge distillation; (2) Distillation from foundation models; and (3) Semi-supervised feature selection. Figure \ref{fig:ukdsl} shows the components of UKDSL. We describe each of the components in detail in the following sections, and show our results of how they mitigate the cross-stage miscalibration and improve model performance.

\begin{figure}[t]
  \centering
   \includegraphics[width=8.5cm]{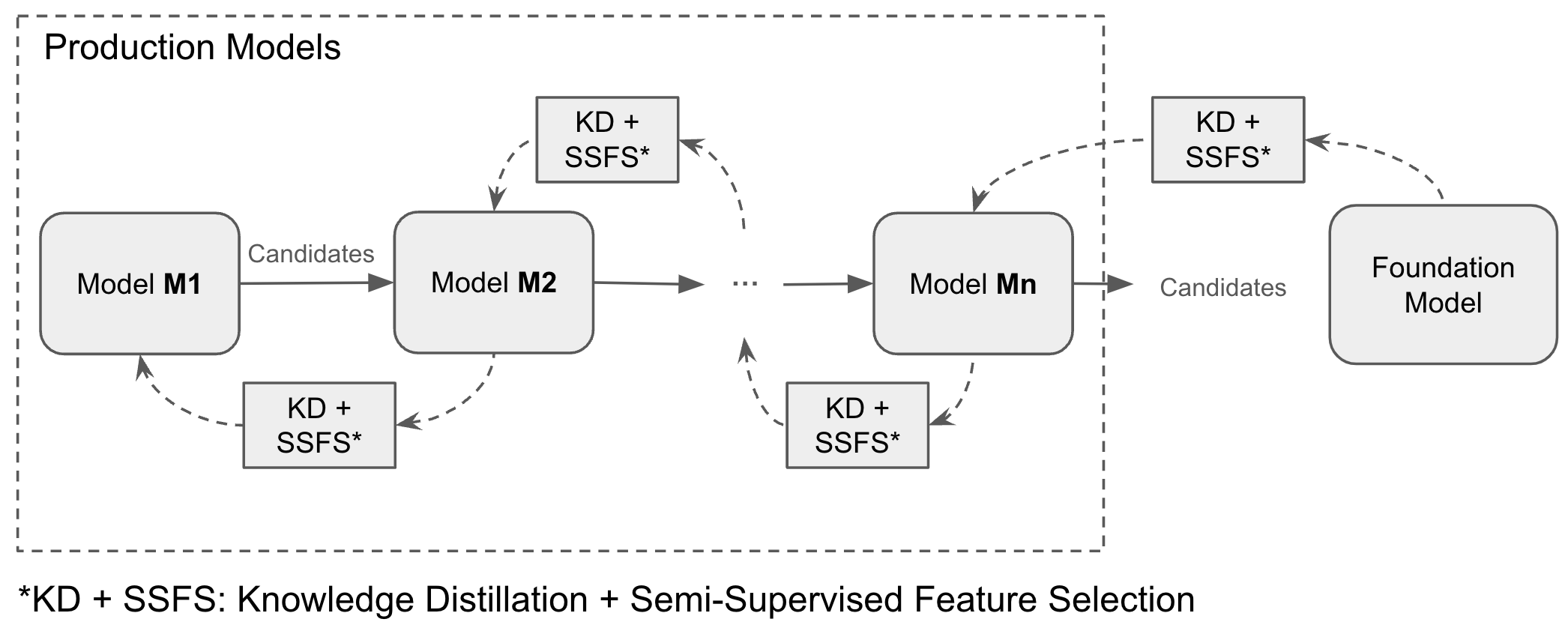}
  \caption{UKDSL: Unified Framework for Knowledge Distillation and Semi-Supervised Learning}
  \label{fig:ukdsl}
\end{figure}

\subsection{Semi-Supervised Cross-Stage Knowledge Distillation}
\label{subsec:semi_consistency_calib}

The typical knowledge distillation setup involves a teacher model and a student model. The student model is the model that is used to serve production traffic, which is also the model that we aim at improving. The teacher model is usually a separate model or an ensemble that's more complex and more accurate. Due to the scale of data and variety of models, it is difficulty and cost inefficient to develop and maintain separate dedicated teachers for all models in an ads delivery system. Within the framework of UKDSL, we solve this problem by semi-supervised cross-stage distillation, where we used later stage models to teach earlier stage models on both labeled and unlabeled data.

Suppose an ads delivery system consists of $N$ stages. Each stage $j$ has a model $\mathbf{M_j}$, which ranks and selects the top candidates from its candidate pool. $\sS_j$ is the set of candidates selected by $\mathbf{M_j}$. $\sS_0$ is the set of all ads candidates before any ranking and filtering. 
Further suppose that this cascade design~\cite{wang2023towards} in multi-stage ranking systems, which is a widely used design pattern in the industry, early stage models need to rank a large pool of candidates (potentially in the order of billions), hence they need to be fast and relatively simple. 
The later stage models only need to rank the selected top candidates by the previous stage. Depending on the stage, the pool size could be in the order of 10k or even hundreds. Therefore they can afford to be slower. As a result, the later stage models usually are larger, more complex, trained using more data/features, thus performing a lot better than earlier stage models. In our notation,  $\mathbf{M_j}$ is one stage after  $\mathbf{M_{j-1}}$.
Consequently, the following properties hold:

\begin{enumerate}
    \item \label{cali-property} If both $\mathbf{M_{j-1}}$ and $\mathbf{M_j}$ are unbiased, then $\mathbf{M_j}$ is well calibrated on $\sS_{j-1}$, whereas $\mathbf{M_{j-1}}$ is mis-calibrated on $\sS_{j-1}$ due to the inherent miscalibration discussed in Section \ref{subsec:inherent_miscalibration}.
    \item \label{accu-property} $\mathbf{M_i}$ is more accurate than $\mathbf{M_j}$ if $i > j$
    \item \label{data-availability-property} $\mathbf{M_j}$ is used in production to serve $\sS_j$, thus all features required by $\mathbf{M_j}$ and its predictions are available for $\sS_j$ from production logging
    \item \label{label-property} $\vx \in \sS_0$ is labeled if and only if $\vx \in \sS_N$
    \item \label{set-property} $\sS_N \subset \sS_{N-1} \subset \dots \subset \sS_0$
\end{enumerate}

Due to (\ref{cali-property}) and (\ref{accu-property}), $\mathbf{M_j}$ is a good teacher model for $\mathbf{M_{j-1}}$ for all $j \in [2, N]$. Further, due to (\ref{data-availability-property}), (\ref{label-property}) and (\ref{set-property}), $\mathbf{M_j}$ can be used to impute pseudo labels for $\mathbf{M_{j-1}}$ on unlabeled data that is much larger than the labeled data, thus harvesting the benefits of semi-supervised learning. Remarkably, (\ref{data-availability-property}) means development and maintenance of new teacher models is not required, as the teacher predictions are readily available in the set they are expected to function as teachers. This not only enables knowledge distillation in an extremely cost effective way, but also overcomes the scale and diversity of model types and ranking stages. Figure \ref{fig:cross-stage-distillation} illustrates the semi-supervised cross-stage distillation with an example 3-stage system.

\begin{figure}[t]
  \centering
   \includegraphics[width=8.5cm]{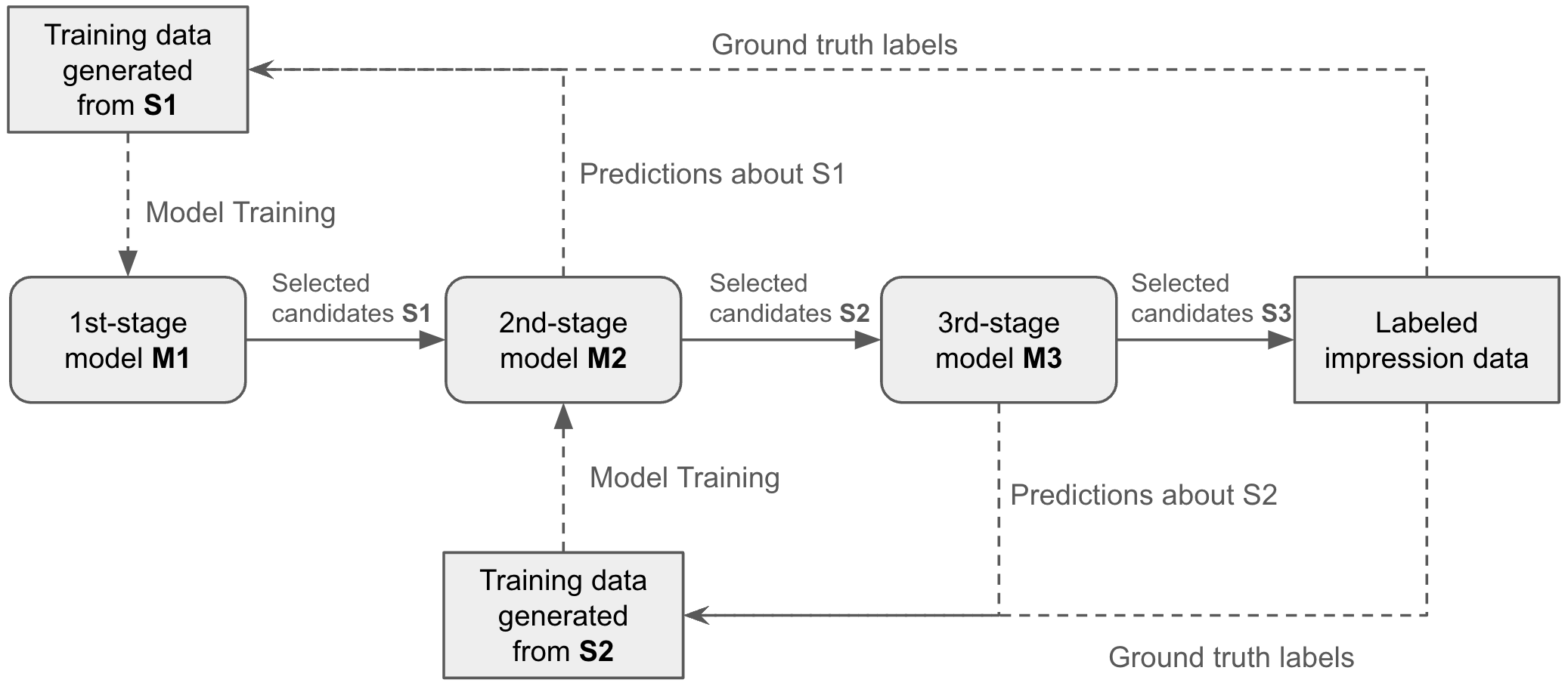}
  \caption{Semi-Supervised Cross-Stage Distillation illustrated with a 3-stage system: each model acts as the teacher for the previous stage on both labeled and unlabeled data}
  \label{fig:cross-stage-distillation}
\end{figure}

\subsubsection{Semi-Supervised Knowledge Distillation}
We make use of the standard binary cross entropy loss for cross-stage distillation. Typical ads ranking models such as CTR and CVR models produces a single probability value denoted by $\vz$, and the models are trained using binary cross entropy loss $\mathcal{L_{\text{BCE}}}(\vy,\vz)$ where $\vy$ is the ground truth label:

\begin{equation}
\label{eq:bce}
    \mathcal{L_{\text{BCE}}}(\vy,\vz) = -\left[\vy  \log({\vz}) + (1 - \vy)  \log{(1 - \vz)}\right]
\end{equation}
In our cross-stage distillation setup, instead of relying on the ground truth labels $\vy$, we rely on the predictions made by a teacher model $\vz_T$ as pseudo labels. The student model is similarly trained using binary cross entropy loss $ \mathcal{L_{\text{BCE}}}(\vz_T,\vz)$. 

\subsubsection{Performance of Semi-Supervised Cross-Stage Distillation}
In table \ref{cross_stage_cd_distill}, we present our model performance results after applying cross-stage distillation. We use the following two metrics for performance measurement. 
\newline
\begin{enumerate}
    \item \textbf{Calibration}: $\sum{\hat{p}_{pre-ranking}}/\sum{\hat{p}_{ranking}}$. Calibration is defined as the ratio of the average predictions of the test model over the average predictions of a reference model or the ground truths. A value close to 1 indicates consistency across two ranking stages when using a reference model, or unbiasedness when using ground truths.
    \item \textbf{Normalized Entropy (NE)}: NE is defined as:\\ $-\sum_{i,c}{\vy_{i,c}\log({p_{i,c}})}/-\sum_{i,c}{\vy_{i,c}\log({\bar{\vy_c}})}$, where $\vy_{i,c}$ is the label for the $i^{th}$ data on class $c$, and $p_{i,c}$ is the corresponding prediction. $\bar{\vy_c}$ is the average probability for class $c$ being positive. The smaller NE suggest better model performance. In the case of consideration data, where there is no ground truth label, as they were not sent to users, we use the later stage model's prediction as the label to calculate NE as a measure of cross stage consistency. 
\end{enumerate}

\vspace{6pt}

The results show that on impression data, the average predictions from pre-ranking model and late stage ranking model are very close to each other. However, pre-ranking model generates higher prediction values than late stage ranking model on consideration data which aligns with the theoretical analysis done in Section ~\ref{subsec:inherent_miscalibration}. Given consideration data represents the actual serving traffic of our models, improving the consistency on consideration data traffic would be crucial for our system performance. By following the analysis done in Section ~\ref{subsec:inherent_miscalibration} and adding cross stage co-training and distillation on consideration data, we allow late stage models generate predictions as supervision for pre-ranking models on consideration data on the fly. This approach improved cross stage consistency by reducing cross stage mis-calibration and reduce cross stage NE, with minimal impact on performances on impression traffic. 

\begin{table}[t]
    \caption{Offline performance comparison between model trained with cross stage distillation on consideration data and baseline.
    NE relative change: lower is better ($\downarrow$).
    Calibration: 1.0 is best.
    }
    \label{cross_stage_cd_distill}
    \centering
    \begin{tabular}{ccc}
    \toprule
     Data & Baseline &  w/ Distillation\\
    \toprule
     &\multicolumn{2}{c}{Model calibration}\\
    \midrule
    Impression Data & 1.01 & 1.03 \\
    \hline
    Consideration Data & 1.27 & 1.12 \\
    \toprule
     &\multicolumn{2}{c}{NE relative change ($\downarrow$)}\\
    \midrule
    Impression Data & 0\% & 0.01\%  \\
    \hline
    Consideration data & 0\% & -1.98\%  \\
    \bottomrule
    \end{tabular}
\end{table}

\subsection{Semi-Supervised Feature Selection (SSFS)}
Selection bias in training data can inadvertently lead to the choice of features that aren't necessarily the most beneficial for the serving data. 
This challenge highlights the necessity for a robust feature selection approach, specifically optimized for distillation, that can effectively reconcile the differences between training and serving data. This forms the motivation for a novel feature selection methodology.

Our approach is illustrated in Figure~\ref{fig:cdfaf}, which consists of following main steps:
\newline
\begin{enumerate}
    \item \textbf{Building the Teacher Model:} We begin by training a teacher model on the impression dataset. This teacher model is then used to generate labels for the unlabeled consideration dataset.
    \item \textbf{Label Augmentation:} Next, we augment the unlabeled dataset with the labels generated by the teacher model.
    \item \textbf{Training a Simplified Model:} Subsequently, we train a simpler version of the student model using all candidate features on a mixed dataset with both labeled and unlabeld data. 
    \item \textbf{Perturbation-based Features Importance:} Then, we compute the feature importance scores for each feature in a withheld dataset using a perturbation-based feature importance algorithm with the simplified model.
    \item \textbf{Combine biased and unbiased features:} Finally, we combine the biased features ranked using regular approach and the unbiased features ranked using our approach based on their relative rank and/or weighted feature importance, and select the best mix that balances the results on both the impression and consideration data sets.
\end{enumerate}
\vspace{6pt}
A high level pseudo-code algorithm for perturbation-based feature importance is given below:

\begin{algorithm}[h]
\caption{Perturbation-based feature importance}\label{alg:fi}
\begin{algorithmic}
\For{every batch of data}
    \For{every feature}
        \State 1. Measure the original loss with feature
        \State 2. Random shuffle feature values in the batch
        \State 3. Measure the loss on the shuffled batch
        \State 4. Report feature importance as loss change
    \EndFor
\EndFor
\State Report mean \& stdev for feature importance
\end{algorithmic}
\end{algorithm}

\begin{figure}[t]
\centering
\hspace*{0.5cm} 
\includegraphics[width=8.0cm]{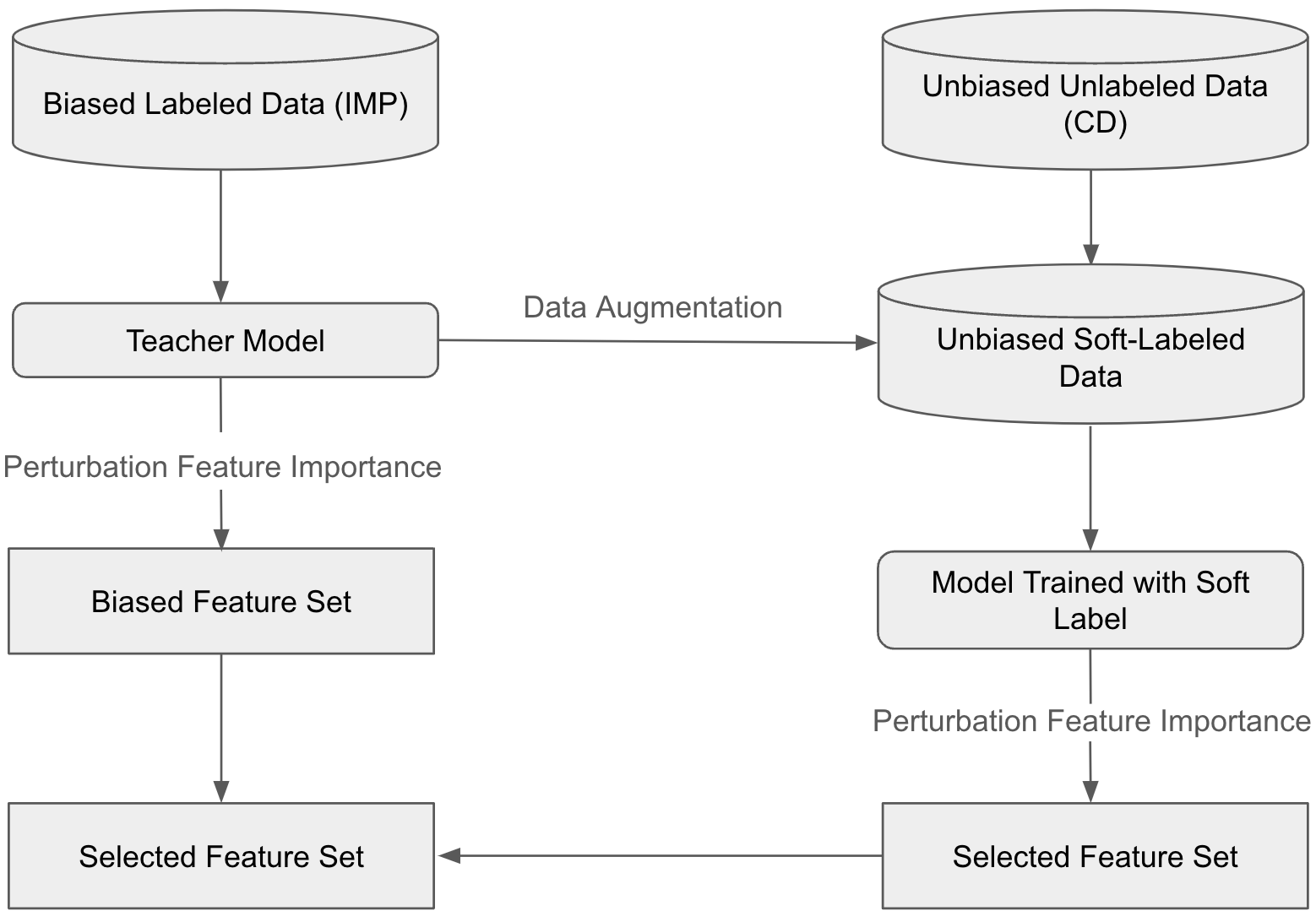}
\caption{This diagram illustrates the Semi-Supervised Feature Selection module used by UKDSL. A set of unbiased features is selected and combined with the regular biased features to create the final set of features for use in the model.}
\label{fig:cdfaf}
\end{figure}

Table~\ref{tb:cdfaf} compares different methods for combining biased and unbiased features in terms of NE improvement on the \emph{impression data set (IMP)} and \emph{consideration data set (CD)}. 
CD Importance Only leads to a 0.023\% gain in IMP NE and a 0.079\% gain in CD NE. Combining the CD Feature Importance with IMP Feature Importance by their relative rank in feature importance improves CD NE further but negatively impacts IMP NE. The best method, Union of Top Features, which selects top features using IMP feature importance and CD feature importance independently and take the union of the two sets, led to 0.137\% gain in IMP NE and 0.725\% gain in CD NE, proving its superior generalizability in consideration data.

\begin{table}[t]
\caption{Comparison of different methods for combining biased and unbiased features in terms of relative NE change (lower is better ($\downarrow$)) on the impression data and consideration data.}
\label{tb:cdfaf}
\centering
\begin{tabular}{lcc}
\toprule
& \multicolumn{2}{c}{NE relatvie change  ($\downarrow$)} \\
\midrule
Method & Impression Data& Consideration Data\\
\midrule
IMP Importance Only & 0.000\% & 0.000\% \\
\hline
CD Importance Only & -0.023\%& -0.079\%\\
\hline
Average Rank & 0.209\%& -0.371\%\\
\hline
Average Importance & -0.009\%& -0.094\%\\
\hline
Intersection of Top Features & 0.018\%& -0.079\%\\
\hline
\textbf{Union of Top Features} & -0.137\%& -0.725\%\\
\bottomrule
\end{tabular}
\end{table}

\subsection{Semi-Supervised Learning from Foundation Models (SSLFM)}
In our semi-supervised cross-stage distillation described in Section \ref{subsec:semi_consistency_calib}, we always use the next-stage model as the teacher. However, this approach does not work for the model in the last stage, for which no production model can be used as a teacher. With the recent rise in popularity of foundation models, we leverage foundation models as the teacher models. Unlike production models, the foundation models are not used to serve production traffic. Therefore, they are not subject to the same capacity constraint as the production models. This allows them to have increased model complexity, consume more features, and make use of multiple related tasks with a multi-task learning setup to gain additional information during training. As a result, the quality of predictions by the foundation models are far superior than the production models thus can be used as teacher labels for the production models.

In our experiments, we found that using multi-task learning helps improve the knowledge transfer efficiency when foundation models are used as teachers. Specifically, we add an auxiliary task and a dependent task in addition to the main task (e.g. CTR task) in a multi-task learning manner. This approach is illustrated in Figure \ref{fig:ssl_em_fig}.
\newline
\begin{enumerate}
    \item \textbf{Dependent task}: On top of the main task (e.g. CTR prediction), we add a dependent task that takes the output of the CTR task as input. This dependent task is trained on unlabeled data only, with the foundation model's output as teacher label.
    \item \textbf{Auxiliary task}: We add another auxiliary task that takes the shared feature representation as input. Like the dependent task, this auxiliary task is trained on unlabeled data only, with the foundation model's output as teacher label.
\end{enumerate}
\vspace{6pt}
\begin{figure}[t]
  \centering
   \includegraphics[width=8cm]{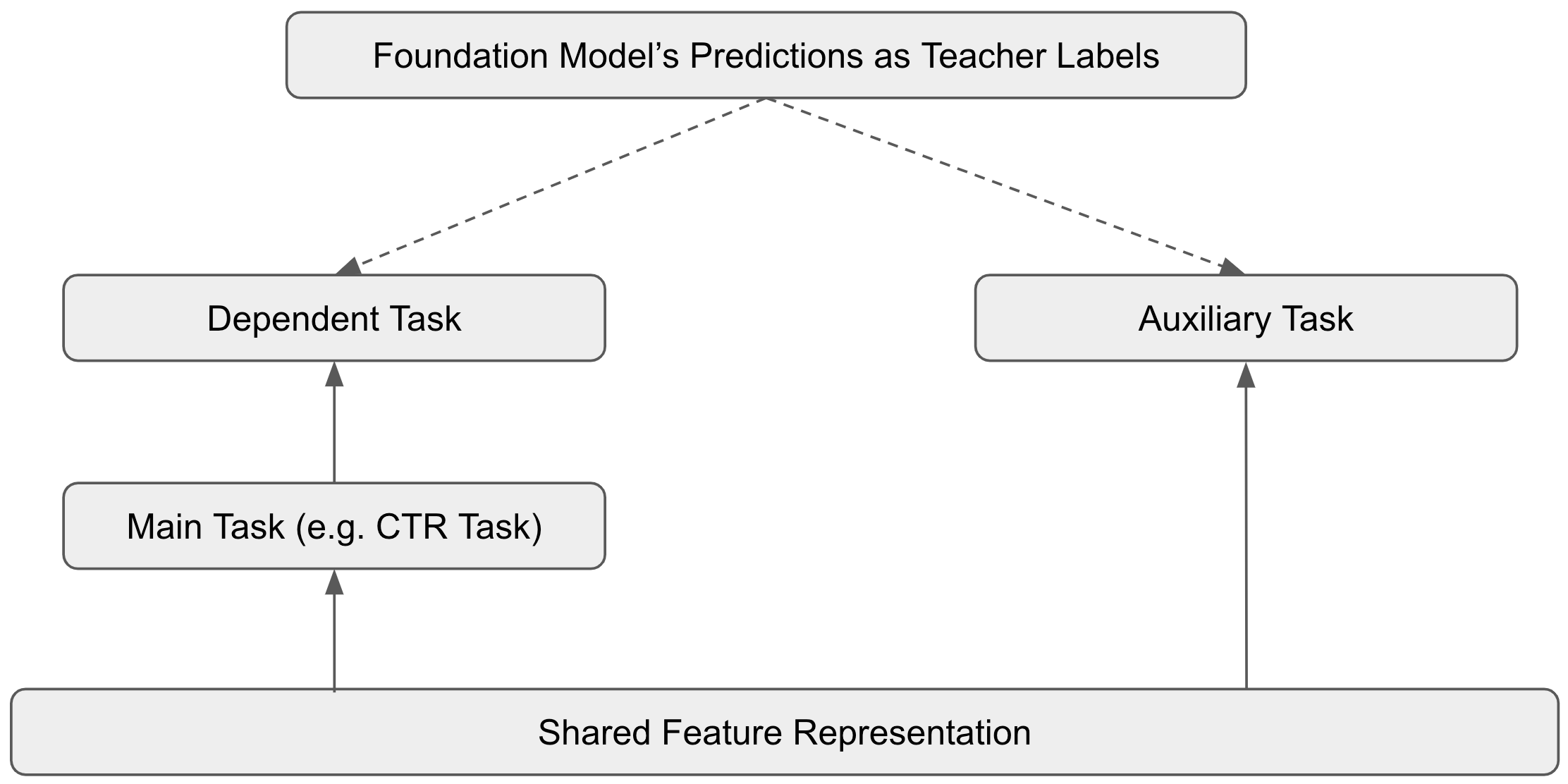}
  \caption{Diagram of knowledge distillation implementation in the SSLFM module for an example CTR prediction model. We add an auxiliary task and a dependent task to the student model in a multi-task setting. Both tasks are trained to predict the teacher model's predictions}.
\label{fig:ssl_em_fig}
\end{figure}

Table \ref{ssl-em} summarizes our experiment results for a CTR model. Notably, we found our semi-supervised learning implementation with foundation models significantly improved the student model's NE on the impression data.

\begin{table}[t]
\caption{Performance comparison between different variants of SSLEM models and baseline: NE relative change (lower the better$\downarrow$)}
\label{ssl-em}
\centering
\begin{tabular}{lcc}
\toprule
 Model & NE relative change ($\downarrow$)   \\
\midrule
Baseline & 0.000\%\\
\hline
Dependent task only & -0.151\%  \\
\hline
Auxiliary task only & -0.159\%  \\
\hline
Dependent + auxiliary task & -0.163\% \\
\bottomrule
\end{tabular}
\end{table}

\subsection{Baselines}

The experiment comparisons in this manuscript are all compared against the latest production models in a multi-billion-scale industrial ads ranking system, prior to the adoption of UKDSL. 
Our criteria for selecting baselines was to identify models that 1) have been proven to operate effectively at the industry scale; 2) represent the  state-of-the-art ads ranking product models  in the industry.
We consider these production recommendation models to be among the state-of-the-art baselines that meet the above criterion.

\section{Conclusion and Future Work}
\label{sec:discussions_future}

One big drawback of the traditional ads ranking approaches that rely solely on supervised learning is their inherent inability to generalize to the ad candidates that have not yet been delivered to users for ad impressions. Semi-supervised learning can be used to mitigate this drawback, and our proposed UKDSL framework enables the application of semi-supervised learning in an industrial-grade ads ranking systems at scale, allowing developers to deploy semi-supervised learning to all model types across all stages with minimal additional cost and effort. Our experiment results show that the UKDSL framework helps overcome the inherent mis-calibration issue, thus improving model calibrations, but also improves their performance on both labeled and unlabeled data.
UKDSL has been launched to major industrial-scale ads recommendation models across different ranking stages and traffic. This indicates that it can be generalized to diverse user demographics and content types, considering the scale and reach of the deployed ads platform.

By sharing our success stories in leveraging unlabeled data in an industrial-scale ads ranking systems, we hope to motivate more future work in this important and impactful area of research. Particularly, the use of foundation models for distillation on earlier stage models is an interesting and currently under-explored direction. Having the same foundation model as the teacher for all ranking stages can not only potentially improve their individual model performance, but also further improve the cross-stage prediction consistency. Other areas worth further exploring include developing more efficient foundation models, improving knowledge transfer efficiency, and improved sampling methods for learning from unlabeled data more efficiently. 

\section*{Acknowledgment}
We thank Yi-Hsuan Hsieh and Song Zhou for building the infrastructure that enabled the usage of consideration data in our system.

\clearpage
\newpage
\bibliographystyle{assets/plainnat}
\bibliography{paper}

\clearpage

\end{document}